# InfantFace: Detecting infant faces in neonatal clinical environments

Abdullah Bin-Obaid[1,*], Maria M. Cobo[2,3], Rebeccah Slater[2], Lionel Tarassenko[1], and Mauricio Villarroel[1,*]

[1]Institute of Biomedical Engineering, Department of Engineering Science, University of Oxford, Oxford, UK

[2]Department of Paediatrics, University of Oxford, Oxford, UK

[3]Universidad San Francisco de Quito USFQ, Colegio de Ciencias Biologicas y Ambientales, Quito, Ecuador

---

***Corresponding authors:**

Abdullah Bin-Obaid, abdullah.binobaid@eng.ox.ac.uk;

Mauricio Villarroel, mauricio.villarroel@eng.ox.ac.uk

## ABSTRACT

Reliable localisation of the neonatal face is the first step for several video-camera based non-contact assessments such as pain and distress related facial expression analysis, pain scoring, cardiorespiratory signal extraction and cessation of breathing alerts. However, major challenges persist in neonatal clinical environments. Cluttered backgrounds, illumination changes and poor lighting conditions can reduce the accuracy of face detection models. Clinical interventions, monitoring equipment and, in some cases, medical devices can obstruct the face, making visual assessment difficult.

We propose a one-stage YOLOv11m-based model tailored for face detection of infants in neonatal clinical environments. We combined multiple publicly available datasets (VGGFace2, CelebA, FDDB, WIDER FACE) to train and evaluate our proposed model. We then fine-tuned our model on a neonatal research dataset involving 228 videos from 114 recording sessions of 113 independent infants. Before fine-tuning, our model achieved an $AP_{50}$ of 0.87, surpassing the performance of three state-of-the-art general face detectors. Performance improved further to an $AP_{50}$ of 0.96 after clinical-domain adaptation.

Evaluating face detection performance across different datasets remains a challenge due to the lack of publicly available neonatal datasets. Prioritising the creation of such datasets, while upholding appropriate privacy safeguards and ethical standards in their creation and use, would greatly support further progress in this field.

## 1 Introduction

Detecting a face in an image is a computationally demanding task. Examples of challenges in developing face detection methods include cluttered backgrounds, where many objects are present within the frame; face occlusion, objects covering part of the face; scaling, as faces can appear small or large based on the distance between the camera and the individual; illumination changes, as poor lighting and shadows reduce the accuracy of methods; image orientation from the camera's optical axis; presence of more than one face in an image; and skin colour[1–3].

Improving the generalisation of face detection models to new target domains is a major challenge, particularly for clinical applications. Models trained on one dataset might perform poorly when deployed to other environments due to variations in appearance, pose, image quality and demographics of the target population[4].

Hospital infant-care environments range from postnatal wards, where infants are generally healthy, to neonatal units, where infants may require different levels of clinical support. Medical devices such as feeding tubes, endotracheal tube fixations and phototherapy goggles frequently obstruct the face, making visual assessment difficult. These occlusions, combined with variable lighting and patient fragility, pose challenges for facial analysis[5]. In addition, routine interventions such as blood sampling, feeding, nappy changes and ventilation support are carried out regularly, further contributing to motion and visual variability.

Face detection plays a key role in a range of non-contact assessment and physiological monitoring research in neonatal care. It is used as the first step to develop methods for facial expression analysis to assess pain, distress and discomfort[6–9]. It also serves as the first stage to extract cardio-synchronous photoplethysmographic imaging (PPGi) signals from which heart rate and respiratory rate can be estimated using video cameras[10–13].

In this work, we propose an end-to-end pipeline for face detection of infants in neonatal clinical environments. Our approach employs a You Only Look Once (YOLO) object detection model, initially pre-trained on the Common Objects in Context (COCO) dataset for general object detection. We trained our model on a large-scale face dataset comprised of more than 3 million images with variations in pose, age, illumination and ethnicity. We subsequently fine-tuned our model on three public datasets widely used to benchmark general-purpose face detectors. Finally, we further fine-tuned our model on video recordings of infants undergoing acutely painful clinical procedures in hospital environments. In contrast to previous studies, our study covers the full pipeline, from data preparation and training strategies to evaluation. We used an open-source model, released the annotations used in the training process and reported performance using standard object detection metrics to ensure fair and reproducible evaluation.

## 1.1 Related work

Several prominent deep learning algorithms have been introduced for face detection tasks. Multi-Task Cascaded Convolutional Neural Networks (MTCNN)[14] employs a cascaded, three-stage architecture that progressively refines face candidates by combining face detection and alignment in a single, end-to-end framework. Although now outperformed by more recent models, MTCNN remains a widely used baseline due to its simplicity.

BlazeFace[15] was developed by Google as part of the MediaPipe library. It is a lightweight model designed specifically for mobile devices and capable of running in real time. BlazeFace was inspired by MobileNets[16], which enabled efficient feature extraction for lightweight object detection. It uses a compact, single-shot architecture based on anchor boxes and depthwise separable convolutions, which reduce computation while maintaining accuracy. The model is intended for faces captured within approximately two metres of a smartphone camera. It is not optimised for distant subjects or large variations in pose, which limits its applicability for more general face detection tasks.

RetinaFace[17] is a one-stage face detector built as an extension of RetinaNet[18], originally proposed for general object detection. It was designed for detecting faces in the wild, with a focus on handling variations in scale, pose and occlusion. It employs a Feature Pyramid Network (FPN) to capture multi-scale features and enhances them with context modules that expand the receptive field for better detection of small or occluded faces. RetinaFace remains a widely adopted baseline in many benchmarks and continues to be used in both research and real-world applications.

Sample and Computation Redistribution for Face Detection (SCRFD)[19] is a one-stage face detector proposed with a focus on improving computational efficiency and supporting real-time applications. It introduced two key innovations: a Sample Redistribution Strategy (SRS) and a Computation Redistribution Strategy (CRS). SRS improves small-face detection by increasing training samples in earlier layers of the network. The CRS optimises resource allocation by shrinking the search space and applying a two-step computation-redistribution search. These design choices aim to balance detection accuracy and computational cost.

Dual Shot Face Detector (DSFD)[20] is a two-stage face detector designed to improve detection accuracy of small and occluded faces through a dual-shot structure and enhanced feature representation. It introduces a Feature Enhance Module (FEM) to combine multi-level information and extract richer features by expanding the receptive field. DSFD also improves the design by adjusting anchor ratios based on the distribution of face scales in the training dataset.

These general face detectors were developed using datasets featuring adults, adolescents and young children. Despite their broader applicability, several studies[21-24] highlighted a decrease in their performance when applied to the clinical neonatal environment. The specific challenges of clinical neonatal environments, including the Neonatal Intensive Care Unit (NICU), combined with the limited performance of general face detectors in this context, have motivated the research community to develop models specifically tailored to the unique conditions of clinical neonatal environments.

Dosso et al.[23] evaluated several face detection models on NICU data and subsequently selected two for fine-tuning (RetinaFace and YOLO5Face models) on a dataset consisting of 2,714 images from 317 participants. They adopted a low Intersection over Union (IoU) threshold during training with YOLO5Face (0.2). This can overestimate performance by accepting loosely aligned bounding boxes and reduce localisation accuracy. At evaluation, Dosso et al. selected only the highest-confidence detection per image, so extra false positives in the frame were not penalised, which can overstate precision in cluttered NICU scenes.

Hausmann et al.[7] proposed YOLOv5- and YOLOv6-based face detection models trained on 8,826 images from 36 infants and tested on 12,416 images from 9 infants. As a baseline, they used an off-the-shelf general object detector. This baseline model was compared against models specifically trained for face detection, limiting the validity of the performance comparison. To assess generalisability, they evaluated their model on a separate dataset from 14 additional infants. Accuracy decreased from 99% to 62.7%, suggesting possible limitations in generalisability across datasets.

Gleichauf et al.[24] proposed a detection framework that uses Red Green Blue (RGB) and thermal images. They trained RetinaNet and YOLOv3 on a dataset consisting of 1,700 images from four infants (three full-term and one pre-term) and tested on 65 images from one full-term infant. While the sensor fusion approach is well described in the literature on neonatal monitoring, the study was conducted on a limited dataset. Their proposed methods were also developed to detect the head region, not the face.

## 2 Results

### 2.1 Evaluation metrics

We report $AP_{50}$, the Average Precision (AP) evaluated at an IoU threshold of 0.5; and $AP_{50:95}$, the mean of AP values computed at different IoU thresholds ranging from 0.50 to 0.95 in increments of 0.05. The AP is defined as:

$$\mathrm{AP} = \sum_{n} (R_n - R_{n-1}) \cdot P_n \quad (1)$$

where $R_n$ and $P_n$ are the recall and precision at the $n^{th}$ point on the precision-recall curve, respectively.

The inclusion of the $AP_{50:95}$ metric provides a stricter and more informative measure of localisation precision. When a model achieves high $AP_{50}$ but low $AP_{50:95}$, this indicates that the model is able to find a face, but does not localise the bounding box with high accuracy.

### 2.2 Datasets

Table 1 summarises the datasets used for training and testing. VGGFace2 served as the base training dataset, while CelebA, FDDB and WIDER FACE were used for fine-tuning. The neonatal research dataset was introduced for domain adaptation to ensure relevance to the target application.

**Table 1. Summary of the face datasets used for the development of our proposed method.**

| Dataset | Total | | Training | | Test | |
|---|---|---|---|---|---|---|
| | Images | Subjects | Images | Subjects | Images | Subjects |
| **Base training dataset** | | | | | | |
| VGGFace2 | 2,758,623 | 8,691 | 2,206,898 | 6,950 | 551,725 | 1,741 |
| **Fine-tuning datasets** | | | | | | |
| CelebA | 191,214 | 10,174 | 153,218 | 8,139 | 37,996 | 2,035 |
| FDDB | 1,608 | – | 1,287 | – | 321 | – |
| WIDER FACE | 4,765 | – | 3,780 | – | 985 | – |
| **Domain adaptation** | | | | | | |
| Neonatal research | 2,052 | 113 | 1,026 | 56 | 1,026 | 57 |
| **Total** | 2,958,262 | 18,978 | 2,366,209 | 15,146 | 592,053 | 3,833 |

## 2.3 Performance of general face detectors

Table 2 reports the performance of the three general face detectors MTCNN[14], SCRFD[19] and DSFD[20] and of the InfantFace model (before fine-tuning) on the test sets shown in figure 1.

**Table 2. Comparison of face detection performance before fine-tuning. Bold indicates the best score per row.**

| Test set | MTCNN[14] (off-the-shelf) | | SCRFD[19] (off-the-shelf) | | DSFD[20] (off-the-shelf) | | InfantFace (pre-domain adaptation) | |
|---|---|---|---|---|---|---|---|---|
| | $AP_{50}$ | $AP_{50:95}$ | $AP_{50}$ | $AP_{50:95}$ | $AP_{50}$ | $AP_{50:95}$ | $AP_{50}$ | $AP_{50:95}$ |
| VGGFace2 | 0.98 | 0.82 | 0.99 | 0.89 | 0.99 | 0.90 | **0.99** | **0.92** |
| CelebA | 0.77 | 0.17 | 0.87 | 0.23 | 0.82 | 0.20 | **0.96** | **0.34** |
| FDDB | 0.99 | 0.39 | **1.00** | 0.51 | **1.00** | 0.48 | 0.99 | **0.61** |
| WIDER FACE | 0.98 | 0.62 | **1.00** | **0.76** | 0.99 | **0.76** | 0.80 | 0.57 |
| Neonatal research | 0.54 | 0.22 | 0.81 | 0.36 | 0.85 | **0.40** | **0.87** | **0.40** |

The InfantFace model obtained the highest $AP_{50}$ on VGGFace2 and CelebA, with its best performance on VGGFace2, an $AP_{50}$ of 0.99 and an $AP_{50:95}$ of 0.92. In contrast, performance on WIDER FACE was the worst, with an $AP_{50}$ of 0.80. The lowest $AP_{50:95}$ of 0.34 was observed on CelebA. Across all models, $AP_{50:95}$ scores were consistently lower than $AP_{50}$, with CelebA showing the largest gap with an $AP_{50}$ of 0.96 and an $AP_{50:95}$ of 0.34. On the neonatal research dataset, all detectors achieved a lower $AP_{50}$ than the InfantFace model, which had an $AP_{50}$ of 0.87.

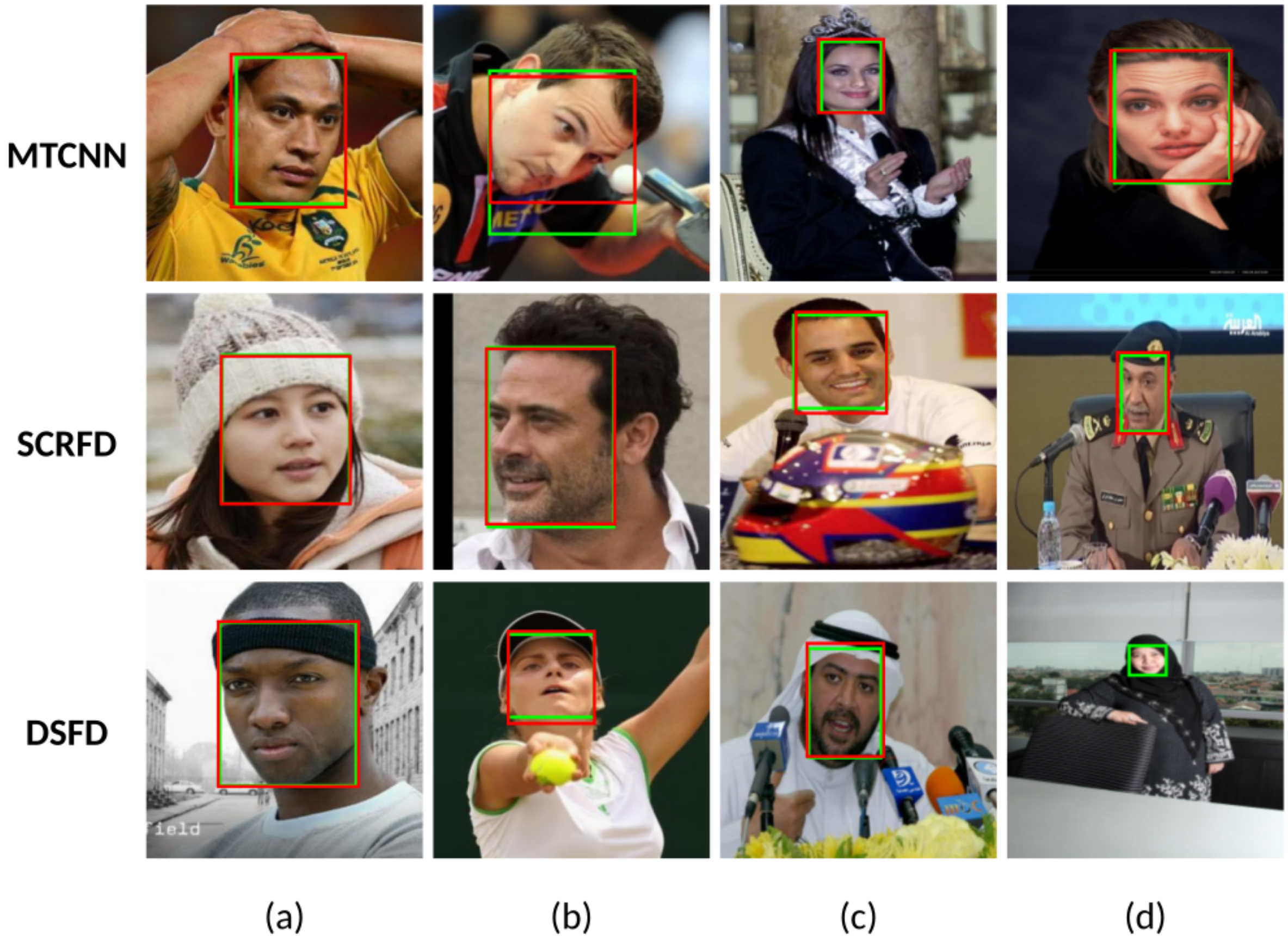

**Figure 1.** Example detections from the evaluated face detectors across public datasets. Columns correspond to (a) VGGFace2, (b) CelebA, (c) FDDB and (d) WIDER FACE. The green bounding boxes correspond to the detections from the face detectors shown in the left column (MTCNN, SCRFD, DSFD). The red bounding boxes correspond to detections from our InfantFace model before domain adaptation. In (d), our model did not detect the bottom image. Images have been slightly cropped and resized to improve visual alignment and ensure consistent presentation.

### 2.4 Performance of fine-tuned models

We fine-tuned the InfantFace model on the training subsets of CelebA, FDDB and WIDER FACE separately. We then evaluated each resulting model across all test sets as shown in table 3.

On the VGGFace2 dataset, the FDDB-tuned and WIDER FACE-tuned models each achieved an $AP_{50}$ of 0.99, with WIDER FACE-tuned obtaining the highest $AP_{50:95}$, with a value of 0.90. On the CelebA dataset, the CelebA-tuned model achieved the best performance with an $AP_{50}$ of 0.99 and an $AP_{50:95}$ of 0.88. On the FDDB dataset, all models achieved an $AP_{50}$ of 0.99, while the FDDB-tuned model led in $AP_{50:95}$, with a value of 0.89. On the WIDER FACE dataset, the WIDER FACE-tuned model obtained the top result with an $AP_{50}$ of 0.92 and an $AP_{50:95}$ of 0.72. On the neonatal research dataset, the WIDER FACE-tuned model achieved the highest scores, with an $AP_{50}$ of 0.90 and an $AP_{50:95}$ of 0.47.

**Table 3. Comparison of InfantFace model performance (before domain adaptation) after fine-tuning on CelebA, FDDB, and WIDER FACE datasets separately. Bold indicates the best score per row.**

| Test set | CelebA-tuned | | FDDB-tuned | | WIDER FACE-tuned | |
|---|---|---|---|---|---|---|
| | $AP_{50}$ | $AP_{50:95}$ | $AP_{50}$ | $AP_{50:95}$ | $AP_{50}$ | $AP_{50:95}$ |
| VGGFace2 | 0.79 | 0.19 | **0.99** | 0.46 | **0.99** | **0.90** |
| CelebA | **0.99** | **0.88** | 0.98 | 0.37 | 0.94 | 0.31 |
| FDDB | **0.99** | 0.35 | **0.99** | **0.89** | **0.99** | 0.57 |
| WIDER FACE | 0.71 | 0.20 | 0.88 | 0.40 | **0.92** | **0.72** |
| Neonatal research | 0.65 | 0.14 | 0.65 | 0.17 | **0.90** | **0.47** |

## 2.5 Model adaptation performance

Table 4 presents the performance of our InfantFace model when fine-tuned on both the WIDER FACE and neonatal research datasets and when performing domain adaptation on the neonatal research dataset alone. The directly adapted model is referred to as InfantFace in the discussion and subsequent sections, unless otherwise stated.

**Table 4. Comparison of InfantFace model variants after neonatal clinical-domain adaptation. Bold indicates the best score per row.**

| Test set | InfantFace (WIDER FACE + neonatal-adapted) | | InfantFace (neonatal-adapted) | |
|---|---|---|---|---|
| | $AP_{50}$ | $AP_{50:95}$ | $AP_{50}$ | $AP_{50:95}$ |
| VGGFace2 | **0.99** | 0.66 | **0.99** | **0.78** |
| CelebA | 0.77 | 0.20 | **0.83** | **0.23** |
| FDDB | **0.99** | 0.39 | **0.99** | **0.44** |
| WIDER FACE | **0.87** | 0.52 | 0.84 | **0.56** |
| Neonatal research | **0.96** | 0.65 | **0.96** | **0.67** |

On the VGGFace2 dataset, both models achieved an $AP_{50}$ of 0.99, with InfantFace having a higher $AP_{50:95}$. On the CelebA dataset, InfantFace obtained the best performance with an $AP_{50}$ of 0.83 and an $AP_{50:95}$ of 0.23. On the FDDB dataset, $AP_{50}$ was 0.99 for both, while InfantFace had the highest $AP_{50:95}$, with a value of 0.44. On the WIDER FACE dataset, InfantFace (WIDER FACE + neonatal-adapted) achieved the highest $AP_{50}$ of 0.87, whereas InfantFace had the highest $AP_{50:95}$ of 0.56. On the neonatal research dataset, both models reached an $AP_{50}$ of 0.96, with InfantFace having a slightly higher $AP_{50:95}$ of 0.67 compared with 0.65.

## 3 Discussion

This study presents a deep-learning, one-stage face detection model. We first trained the model on a large-scale public dataset (VGGFace2) and then fine-tuned it on publicly available benchmark datasets. We then fine-tuned the model on a neonatal research dataset to achieve accurate face detection in challenging neonatal clinical environments. Our training strategy was specifically designed with the clinical neonatal context in mind, for which images typically contain just one infant. In support of reproducibility, we provide the YOLO-based implementation and annotations, whereas some existing models, like DSFD[20], only release pre-trained weights and inference code.

The results presented in table 2 show the performance of our InfantFace model, before fine-tuning, with an $AP_{50}$ of 0.99 on the VGGFace2 and FDDB datasets, but a lower $AP_{50}$ of 0.80 on the WIDER FACE dataset. In contrast, the $AP_{50:95}$ scores varied substantially across datasets. The performance differences between $AP_{50}$ and $AP_{50:95}$ highlight that, whilst a detector is effective at identifying the presence of a face, its bounding box localisation precision remains a key area for improvement. The fact that the highest $AP_{50}$ and $AP_{50:95}$ values were achieved on the VGGFace2 test set is to be expected, as this dataset was used for training the model.

The differences in model performance across datasets, as shown in table 3, highlight the domain-specific characteristics of each dataset. Models fine-tuned on a particular dataset consistently achieved higher performance on its own test split than on other datasets. This suggests that each dataset introduces distinct features or biases to which the model must adapt in order to perform optimally. Further investigation is needed to better understand the trade-offs involved in domain-specific adaptations. In scenarios with limited computational resources, selectively sampling from the public datasets might help to capture a broader range of image characteristics and improve generalisability. Balanced sampling, whereby subsets from larger datasets (e.g., VGGFace2) are downsampled or re-weighted to match the scale of smaller datasets, is an example of such an approach. Sampling can also be performed according to image-level attributes such as pose, occlusion or blur, ensuring a diverse and representative set of facial characteristics. Alternatively, training can use mixed batches with controlled proportions from each dataset to ensure no single dataset dominates.

Table 3 shows that the model fine-tuned on the WIDER FACE dataset generalised better on the neonatal research dataset compared to the models fine-tuned on the CelebA or FDDB datasets. Single-face images extracted from the WIDER FACE dataset share the greatest visual or contextual similarities with the clinical neonatal images. This may be particularly useful in scenarios where clinical neonatal training data are not available, as a model fine-tuned on the WIDER FACE dataset alone was able to achieve a value of 0.90 for $AP_{50}$ using only publicly available datasets.

When comparing the performance of our InfantFace (pre-domain adaptation) model in table 2 with the InfantFace (neonatal-adapted) model in table 4, we observe a decrease in AP values on the fine-tuning datasets, which highlights a trade-off between specialisation to the clinical neonatal domain and retention of performance on general-domain benchmarks, particularly for $AP_{50:95}$. However, WIDER FACE was the least affected. This supports the earlier observation that single-face images from the WIDER FACE dataset share visual and contextual similarities with those in the neonatal research dataset. Although the WIDER FACE-tuned model was the best-performing model on the neonatal research dataset before neonatal adaptation, further fine-tuning yielded InfantFace (WIDER FACE + neonatal-adapted) and improved performance on the neonatal research dataset. However, InfantFace (neonatal-adapted) remained slightly superior in localisation accuracy, as shown by its higher $AP_{50:95}$.

General face detectors underperform in clinical domains, as suggested in[4,21–24] and seen in table 2. Although our InfantFace (pre-domain adaptation) model, which is a general face detector, performed better than the MTCNN, SCRFD and DSFD models before any fine-tuning, it still required domain adaptation to achieve a good performance overall. After fine-tuning on clinical neonatal data, performance improved from an $AP_{50}$ of 0.87 to 0.96 and from an $AP_{50:95}$ of 0.40 to 0.67, as shown in table 4. These results highlight the importance of adapting models to work effectively in the clinical neonatal environment.

A major challenge in this work is the limited availability of neonatal research datasets, as most datasets are not publicly available. This is expected due to privacy and ethical concerns. This constrains our ability to assess the generalisability of the model to other unseen clinical neonatal environments.

To advance the field, future research should focus on developing publicly accessible datasets for training and developing models. This could involve collaborating with hospitals to develop anonymised datasets with adequate consent procedures or by exploring synthetic data generation techniques that simulate neonatal facial characteristics under clinical conditions.

## 4 Methods

### 4.1 Datasets

We use five datasets, four publicly available datasets and a neonatal research dataset. Two of the datasets, FDDB and WIDER FACE, are widely recognised face detection benchmarks:

**VGGFace2**[25] is a large-scale dataset originally compiled for face recognition tasks. It contains 3,163,185 images of 8,691 individuals, collected from Google Image Search. The dataset exhibits substantial variation in age, pose, illumination, profession, ethnicity and background of each subject. All images were captured in unconstrained environments.

**CelebA**[26] is a large-scale dataset containing 202,599 images of 10,177 individuals, with each image featuring a single annotated face. The dataset includes a variety of backgrounds, poses and lighting conditions.

The Face Detection Dataset and Benchmark (**FDDB**)[27] contains 2,845 images and 5,171 annotated faces. It includes images collected from unconstrained settings, such as online news articles. It features a wide range of challenges, including occlusion, pose variation and out-of-focus faces.

**WIDER FACE**[28] is a benchmark dataset containing 32,203 images with annotations for 393,703 faces, exhibiting a high degree of variability in scale, pose and occlusion. The dataset includes images from 61 different event categories, such as festival, sports fan and family groups (see supplementary materials A). Only 50% of the dataset, totalling 16,106 images and 199,128 faces, is publicly available. This dataset is considered to be a key benchmark in the field of face detection.

**The neonatal research dataset** comprises 228 video recordings (114 recording sessions involving 113 independent infants). Each recording session included two 45-second videos, one recording of the clinically-required heel lance procedure, and one paired control recording involving a non-painful stimulus.

These recordings were collected during the Petal trial[29,30] (approved by the London-South East Research Ethics Committee, reference 21/LO/0523) and the ongoing study 'Investigating Pain in the Developing Human Brain' (South Central - Oxford C Research Ethics Committee, reference 12/SC/0447). Written informed parental consent was obtained for all neonates prior to participating, and the studies conformed to the Declaration of Helsinki and Good Clinical Practice standards.

Data were collected between 2012 and 2023 at the John Radcliffe Hospital (Oxford University Hospitals NHS Foundation Trust, UK) and the Royal Devon and Exeter Hospital (Royal Devon University Healthcare NHS Foundation Trust, UK). Postmenstrual age of the neonates ranged from 33 to 43 weeks, with a mean ±SD of 39.3 ±1.9 weeks. All infants underwent clinically-required heel lance procedures, which are routinely used to collect blood samples from an infant's foot for diagnostic testing. The studies took place on the postnatal wards and neonatal units including the Low Dependency Unit (LDU), High Dependency Unit (HDU), and Intensive Therapy Unit (ITU)/NICU. Data collected in these studies included hand-held video recordings to capture behavioural responses and facial expressions to enable pain scoring using the Premature Infant Pain Profile-Revised (PIPP-R) scale[31].

For consistency, we refer to the VGGFace2 dataset as the base training dataset, the CelebA, FDDB and WIDER FACE datasets as the fine-tuning datasets, and the neonatal research dataset as the domain adaptation dataset (see table 1).

### 4.2 Overview of the methodology

We trained a YOLOv11m detector on a large-scale, curated set of single-face images from VGGFace2. The faces were obtained via a consensus of three off-the-shelf detectors (MTCNN [14], SCRFD [19], DSFD [20]). This served as the base training dataset for our InfantFace model (pre-domain adaptation). We also standardised annotations across datasets to a consistent rectangular format for training and subsequent fine-tuning. We trained the model using five-fold cross-validation with subject-separated splits. After this initial training, we fine-tuned separate models on the CelebA, FDDB, WIDER FACE and neonatal research datasets; and evaluated each across all test sets to assess generalisation across datasets and adaptation to the clinical neonatal domain.

### 4.3 Selection of images

The primary motivation for obtaining single face images was the expectation that only a single face would usually be observed in a cot or neonatal incubator. However, there are inconsistencies in the annotations across publicly available datasets. For example, the original bounding box annotations in VGGFace2 were generated using an automated face detector followed by manual verification. Upon manually reviewing a subset of images, we observed that some images contained multiple visible faces, even though only a single face was annotated. A similar issue was identified in the other datasets as well. Examples of these annotation inconsistencies are provided in supplementary materials B.

We employed three state-of-the-art face detection models as annotators to extract single face images. We selected the MTCNN [14], SCRFD [19] and DSFD [20] models, which represent single-stage, two-stage and multi-stage detection approaches, respectively. These models differ in their image processing methodologies, which affect both their complexity and computational speed.

To ensure that an image contained only one face, we ran all three annotators on every candidate image. Any image for which at least one detector failed to return a face was excluded. For each remaining image, we retained the three bounding boxes produced by the detectors and computed their pairwise IoU. Only images for which all pairs had IoU scores of 0.60 or higher were considered valid single face samples. A reference bounding box was then generated by averaging the four corner coordinates from the three models and rounding each coordinate to the nearest pixel.

For the CelebA, FDDB and WIDER FACE datasets, we retained the official ground-truth bounding boxes unchanged. Since these datasets were used for fine-tuning and evaluation, we preserved their original annotations to ensure alignment with established benchmarks. For the VGGFace2 dataset, which served as our base training dataset, we replaced its annotations with the newly computed reference bounding boxes for all selected single-face images.

To create the neonatal research dataset, we extracted one random frame from each 5-second interval of every 45-seconds video, resulting in nine frames per video. To enhance computational efficiency, we down-sampled the frames by a factor of $\frac{1}{3}$ in both dimensions, maintaining the aspect ratio, using bilinear interpolation. The resulting frame sizes were either $426 \times 240$ pixels or $240 \times 426$ pixels, depending on the original aspect ratio of the video. All extracted frames were manually annotated by one of the authors using the Labelme tool[32]. The final dataset comprises 2,052 extracted frames from 114 recording sessions involving 113 independent infants.

The total dataset includes 2,956,210 images from public sources and 2,052 images from the clinical neonatal domain. A summary of the datasets is presented in table 1.

### 4.4 Standardising annotations

To ensure consistency across datasets, we standardised all annotations to a rectangular bounding box format. This step was necessary for compatibility with our training and evaluation pipeline, which expects axis-aligned rectangles. In particular, the FDDB dataset originally provides elliptical annotations, which we converted to bounding boxes by computing the minimal enclosing rectangle for each ellipse (see supplementary materials C).

### 4.5 Model architecture

We designed a training pipeline for neonatal face detection using the YOLOv11m model[33], where "m" denotes the medium-sized version. We employed the YOLOv11m model initially pre-trained on the COCO dataset[34], which provides general-purpose representations for object detection. Since the COCO-pretrained YOLOv11m model is configured for 80 object classes, we reconfigured its detection head by setting the number of output classes to one, corresponding to the single target class "face". The model was then trained using the face annotations, using the COCO-pretrained weights as an initialisation while adapting the detector from general object detection to single-class face detection. The architecture of the YOLOv11m model is illustrated in figure 2.

### 4.6 Model training

For the base training and fine-tuning datasets, we first reserved 20% of the data as a holdout test set. The remaining 80% was then split into 80% for model training and 20% for validation, resulting in an overall distribution of 64% training, 16% validation and 20% testing. Five-fold cross-validation was performed within the remaining 80% of the data by rotating the training and validation splits across folds to obtain robust performance estimates (see table 1).

For datasets where identities were available (VGGFace2 and CelebA) and to prevent data leakage from the same subject appearing in both training and testing sets, we used stratified sampling by identity to maintain a balanced distribution across splits. For the WIDER FACE dataset, we stratified the 4,765 images by category, first allocating 20% to the test set, then splitting the remaining 80% into 80% training and 20% validation. This was applied to each category to ensure consistent representation of challenging scenarios across all subsets. As the FDDB images do not include category or identity metadata for stratification, we randomly split the dataset.

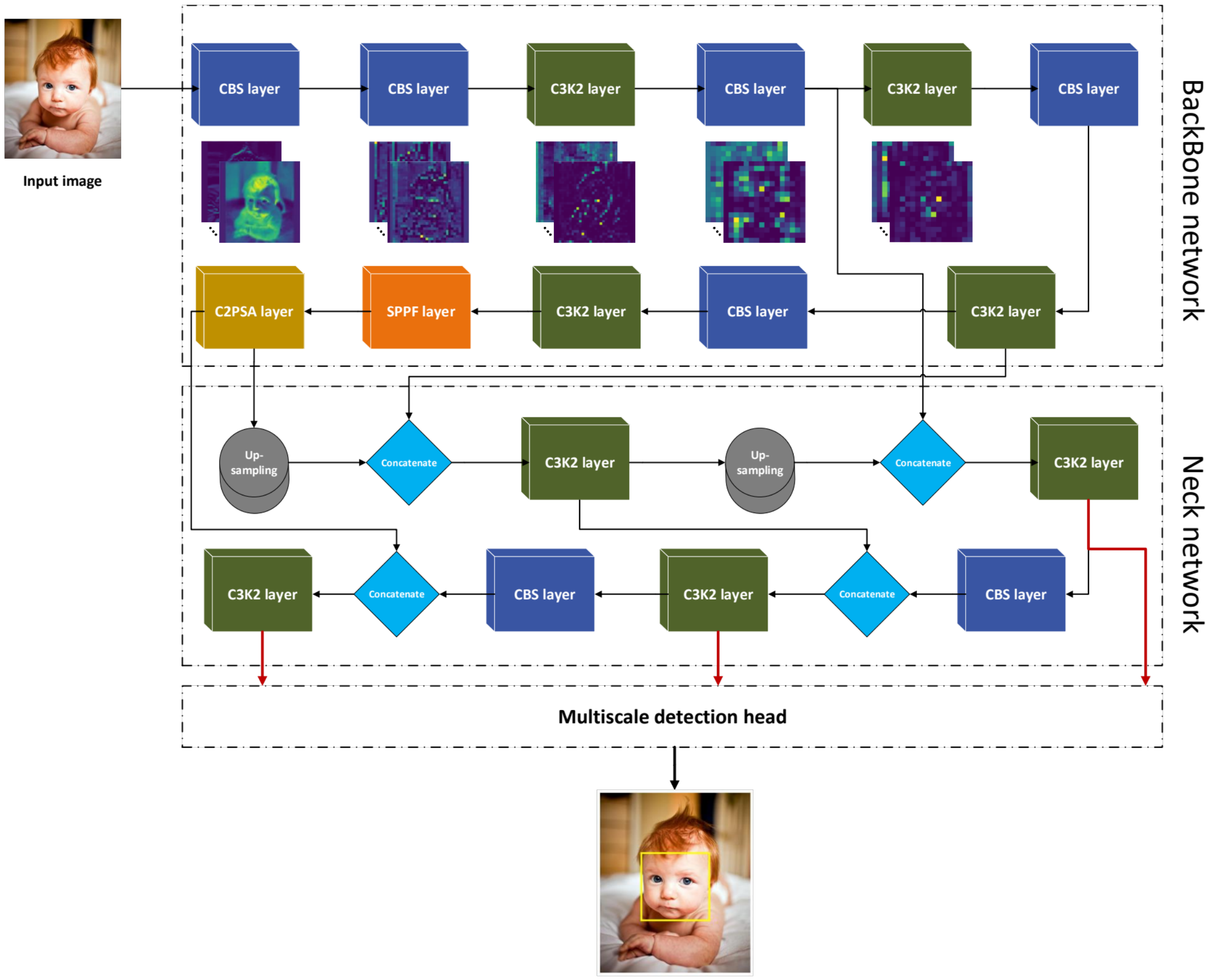

**Figure 2.** Schematic of the YOLOv11m architecture. It highlights the three main components of the network: the backbone, the neck and the multiscale detection head. Functional blocks are colour-coded to distinguish between different operations. The red arrows indicate the flow of multiscale feature maps, extracted at three different spatial resolutions, that are passed from the neck to the detection head. (CBS) stands for Convolution, BatchNorm and SiLU activation and (SPPF) stands for Spatial Pyramid Pooling-Fast. C2PSA and C3K2 are special blocks used in YOLO architecture. The infant image is taken from the publicly available WIDER FACE dataset.

To reduce evaluation bias and account for variance across data splits, we trained the InfantFace model using five-fold cross-validation on the VGGFace2 dataset.

For the neonatal research dataset, we split the data by participant to avoid training and testing on the same infant. The two recording sessions from the same infant were assigned to the same split. We allocated 57 participants to the test set, while the remaining 56 participants were used for model development, including training and validation. This approximately 50/50 split was chosen to maximise the size of the test set, given the relatively small dataset size.

### 4.7 Data augmentation

All images were resized to $128 \times 128$ pixels for training, maintaining aspect ratio by applying zero-padding to the shorter side. Unless otherwise stated, we applied augmentations via the official YOLO pipeline, with default parameters. We applied horizontal flip (left-right mirroring of the image) with a probability of 0.5[35]. Infants in the hospital are placed in a standardised position with their heads oriented toward the head end of the cot or incubator. Therefore, vertical flip was not applied. We applied random rotation within $\pm 90^{\circ}$ to account for tilted heads and occasional camera angle variations[35]. Translation was applied randomly with a 10% shift so that the model could learn to detect partially visible objects (e.g. when an infant's face was partially covered by medical devices). Zero-padding was used to fill the displaced regions. We applied random shear augmentation within $\pm 25^{\circ}$ to introduce variations in appearance due to viewpoint changes, particularly those caused by camera angles relative to the infant's face[35]. We enabled YOLO's perspective augmentation (coefficient of 0.001), which applies a small random projective warp per axis to simulate subtle perspective- and depth-related distortions. To improve the model's ability to detect small faces, such as those of infants at a distance from the camera, we applied multi-scale training with a factor between 0.5 and 1.5 to expose it to a range of face sizes during training. This resulted in the image size varying between $64 \times 64$ and $192 \times 192$ pixels.

We converted all images from RGB to Hue Saturation Value (HSV) colour space. For each image *i*, we computed the mean of each channel $\left(\bar{H}_i, \bar{S}_i, \bar{V}_i\right)$ using OpenCV's scales. We then computed the standard deviation across images of these per-image means to obtain $\sigma_H$, $\sigma_S$ and $\sigma_V$. To set YOLO's HSV augmentation amplitudes, each $\sigma$ was multiplied by 3 (capturing roughly $\pm 3\sigma$ of the colour distribution) and normalised by the channel's numeric range, then clipped to $[0, 1]$. This yielded gains of 0.42, 0.41 and 0.42 for H, S and V, respectively.

### 4.8 Loss function

We defined the total loss as a weighted sum of three components: bounding box loss ($L_{box}$), classification loss ($L_{cls}$) and Distribution Focal Loss (DFL) ($L_{dfl}$), given by:

$$L_{total} = \lambda_{box} L_{box} + \lambda_{cls} L_{cls} + \lambda_{dfl} L_{dfl} \quad (2)$$

where $\lambda_{box}$, $\lambda_{cls}$ and $\lambda_{dfl}$ are the gains for the bounding box, classification and DFL losses respectively. Our dataset consists of only one class (face) images. The face-to-image ratio distribution does not indicate a large number of small faces (see supplementary materials D). We set $\lambda_{box}$ to a value of 8.0 to increase the penalty for inaccurate bounding box predictions and encourage more precise localisation. We reduced the default implemented YOLO gain for $\lambda_{cls}$, which is used for multiple classes (80 classes in COCO), from 0.5 to 0.1 as we have only a single class. We used the default value for $\lambda_{dfl}$ of 1.5 used on the standard YOLO implementation.

Bounding box loss ($L_{box}$) penalises differences between the predicted and ground-truth bounding boxes. It is computed using the Complete-IoU (CIoU), described in[36]. CIoU extends standard IoU by adding penalties for the normalised distance between box centres and for aspect-ratio mismatch. The $L_{box}$ parameter is defined as:

$$L_{box} = 1 - CIoU \quad (3)$$

With

$$CIoU = IoU - \frac{\rho^2(p, p_{gt})}{c^2} - \alpha V$$
$$\alpha = \frac{V}{(1 - IoU) + V} \quad (4)$$
$$V = \frac{4}{\pi^2}\left(\arctan\frac{w}{h} - \arctan\frac{w_{gt}}{h_{gt}}\right)^2$$

where $\rho^2(p, p_{gt})$ is the squared distance between the box centres, $c$ is the diagonal of the smallest enclosing box that covers both predicted and ground-truth boxes and $w, h$ and $w_{gt}, h_{gt}$ are the widths and heights of the predicted and ground-truth boxes, respectively. $V$ is the aspect-ratio penalty and $\alpha$ is the weight for $V$. Both $\alpha$ and $V$ increase the penalty when the IoU is high and the aspect ratio differs.

The classification loss ($L_{cls}$) is defined as the binary cross-entropy for each predicted box:

$$L_{cls} = -\frac{1}{S}\sum_{i=1}^{N}\left[y_i \log \sigma(x_i) + (1 - y_i)\log(1 - \sigma(x_i))\right] \quad (5)$$

where $x_i$ is the predicted logit for the $i$-th candidate box, $y_i \in [0,1]$ is a target indicating how strongly prediction $i$ matches a ground-truth face, $\sigma(\cdot)$ is the sigmoid function, $N$ is the total number of candidate boxes in the batch and $S$ is the normalisation factor defined as the sum of the targets over all candidate boxes.

The DFL loss ($L_{dfl}$) from[37] is defined as:

$$L_{dfl} = -\left[(y_{i+1} - y)\log P_i + (y - y_i)\log P_{i+1}\right] \quad (6)$$

where $y$ is the continuous regression target, $y_i$ and $y_{i+1}$ are the two adjacent discrete values. $P_i$ and $P_{i+1}$ are the predicted probabilities corresponding to $y_i$ and $y_{i+1}$, respectively.

### 4.9 Implementation

We used a batch size of 64. During training, instead of using the general IoU, we used Complete-IoU, which improves localisation accuracy, convergence speed and provides a more informative gradient when predicted and ground-truth boxes overlap but differ in aspect ratio or alignment[36]. We trained the model using the Stochastic Gradient Descent (SGD) optimiser[38] with a learning rate of $10^{-3}$. The update rule used for SGD with momentum was:

$$v_{t+1} = \mu \cdot v_t + g_{t+1}$$
$$p_{t+1} = p_t - \mathrm{lr} \cdot v_{t+1} \quad (7)$$

where $v$ is the velocity, $\mu$ is the momentum coefficient, $g$ is the gradient, $p$ represents the model parameters (i.e., the weights being optimised) and $\mathrm{lr}$ is the learning rate.

For each fold, we trained the model for 50 epochs. Early stopping was applied at the point for which both the validation loss did not decrease and the value of $AP_{50}$ did not increase for 20 consecutive epochs. After completing all five folds, we evaluated the best model from each fold and selected the model from the fold that achieved the highest AP for further fine-tuning and domain adaptation. This model is referred to as the InfantFace (pre-adaptation) model in our work.

After training the InfantFace (pre-adaptation) model, we performed dataset-specific fine-tuning by further training it separately on the training subsets of the CelebA, FDDB and WIDER FACE datasets. We followed the data augmentation and training strategies described in sections 4.7 and 4.9 to fine-tune the InfantFace (pre-adaptation) model. We used a similar approach when we fine-tuned the model on the neonatal research dataset. Since the model had already learned how to detect faces, we set the learning rate to $10^{-4}$ to allow the optimiser to take smaller steps when updating the model parameters and minimising the loss function.

We reduced the batch size to 16, to reflect the relatively smaller size of the fine-tuning dataset. We froze the backbone, which is responsible for extracting general facial feature representations, while keeping the remaining layers unfrozen to allow the model to learn task-specific patterns relevant to the clinical neonatal domain.

During evaluation, we used the largest image size from multi-scale training, $192 \times 192$ pixels and set the confidence threshold for detection at 0.25. This relatively low threshold means that any face predicted with a confidence score above 0.25 is considered a detection. Using a low threshold increases the number of predicted detections, which can lead to more false positives. This evaluation protocol is more challenging for the model, as it requires distinguishing true faces from a greater number of potential false detections, providing a stricter and more realistic assessment of performance in practical scenarios.

## Data availability

The neonatal research dataset used in this study is not publicly available due to its sensitive nature, terms of parental consent, and restrictions imposed by the approved ethics protocol to protect the privacy of the infants involved in the study.

The base training dataset and fine-tuning datasets used are available from their respective original sources and are subject to the terms and conditions of those dataset providers.

## Code availability

Annotations generated from the base training dataset and fine-tuning datasets, training instructions and the corresponding code will be made available at: https://github.com/lcmtlab/InfantFace upon acceptance.

## Acknowledgements

We thank the parents and infants who participated in the original studies from which the neonatal research dataset was derived. The views expressed in this publication are those of the authors and not necessarily those of the NIHR, the University of Oxford, National Health Service, or the UK Department of Health and Social Care.

## Author contributions statement

**M.C.** and **R.S.** collected the data. **R.S.**, **L.T.** and **M.V.** conceptualised the experiment and supervised the project. **A.B.** conducted the experiments, performed the data analysis and wrote the first draft of the manuscript. All authors reviewed the manuscript.

## Funding

**A.B.** acknowledges support from a scholarship awarded by King Saud University. **M.V.** was funded by the Podium Institute for Sports Medicine and Technology, University of Oxford.

## Competing interests

The authors declare no competing interests.

## Consent statement

Written informed parental consent was obtained for all neonates prior to participation.

## References

1. Kumar, A., Kaur, A. & Kumar, M. Face detection techniques: a review. *Artif. Intell. Rev.* 52, 927–948 (2019).

2. Rizvi, Q. M., Agarwal, B. G. & Beg, R. A review on face detection methods. *J. Manag. Dev. Inf. Technol.* 11 (2011).

3. Hasan, M. K., Ahsan, M. S., Newaz, S. S. & Lee, G. M. Human face detection techniques: A comprehensive review and future research directions. *Electronics* 10, 2354 (2021).

4. Huang, B. *et al.* A neonatal dataset and benchmark for non-contact neonatal heart rate monitoring based on spatio-temporal neural networks. *Eng. Appl. Artif. Intell.* 106, 104447 (2021).

5. Heiderich, T. M. *et al.* Face-based automatic pain assessment: challenges and perspectives in neonatal intensive care units. *Jornal de Pediatr.* 99, 546–560 (2023).

6. Bergamasco, L. *et al.* Pain assessment in neonatal clinical practice via facial expression analysis and deep learning. In *International Work-Conference on Bioinformatics and Biomedical Engineering*, 249–263 (Springer, 2024).

7. Hausmann, J. *et al.* Accurate neonatal face detection for improved pain classification in the challenging nicu setting. *IEEE Access* (2024).

8. Ben Aoun, N. Deep learning-based pain intensity estimation from facial expressions. In *International Conference on Intelligent Systems Design and Applications*, 484–493 (Springer, 2023).

9. Giordano, V. *et al.* Comparative analysis of artificial intelligence and expert assessments in detecting neonatal procedural pain. *Sci. Reports* 14, 20374 (2024).

10. Cao, K., Tan, T., Chen, Z., Yang, K. & Sun, Y. A novel heart rate estimation framework with self-correcting face detection for neonatal intensive care unit. *Displays* 85, 102852 (2024).

11. Villarroel, M. *et al.* Continuous non-contact vital sign monitoring in neonatal intensive care unit. *Healthc. technology letters* 1, 87–91 (2014).

12. Chaichulee, S. *et al.* Cardio-respiratory signal extraction from video camera data for continuous non-contact vital sign monitoring using deep learning. *Physiol. measurement* 40, 115001 (2019).

13. Chaichulee, S. *et al.* Localised photoplethysmography imaging for heart rate estimation of pre-term infants in the clinic. In *Optical diagnostics and sensing XVIII: toward point-of-care diagnostics*, vol. 10501, 146–159 (SPIE, 2018).

14. Zhang, K., Zhang, Z., Li, Z. & Qiao, Y. Joint face detection and alignment using multitask cascaded convolutional networks. *IEEE signal processing letters* 23, 1499–1503 (2016).

15. Bazarevsky, V., Kartynnik, Y., Vakunov, A., Raveendran, K. & Grundmann, M. Blazeface: Sub-millisecond neural face detection on mobile gpus. *arXiv preprint arXiv:1907.05047* (2019).

16. Howard, A. G. *et al.* Mobilenets: Efficient convolutional neural networks for mobile vision applications. *arXiv preprint arXiv:1704.04861* (2017).

17. Deng, J. *et al.* Retinaface: Single-stage dense face localisation in the wild. *arXiv preprint arXiv:1905.00641* (2019).

18. Lin, T.-Y., Goyal, P., Girshick, R., He, K. & Dollár, P. Focal loss for dense object detection. In *Proceedings of the IEEE international conference on computer vision*, 2980–2988 (2017).

19. Guo, J., Deng, J., Lattas, A. & Zafeiriou, S. Sample and computation redistribution for efficient face detection. *arXiv preprint arXiv:2105.04714* (2021).

20. Li, J. *et al.* Dsfd: dual shot face detector. In *Proceedings of the IEEE/CVF conference on computer vision and pattern recognition*, 5060–5069 (2019).

21. Dosso, Y. S. *et al.* Neonatal face tracking for non-contact continuous patient monitoring. In *2020 IEEE International Symposium on Medical Measurements and Applications (MeMeA)*, 1–6 (IEEE, 2020).

22. Grooby, E. *et al.* Neonatal face and facial landmark detection from video recordings. In *2023 45th Annual International Conference of the IEEE Engineering in Medicine & Biology Society (EMBC)*, 1–5 (IEEE, 2023).

23. Dosso, Y. S., Kyrollos, D., Greenwood, K. J., Harrold, J. & Green, J. R. Nicuface: Robust neonatal face detection in complex nicu scenes. *IEEE Access* 10, 62893–62909 (2022).

24. Gleichauf, J. *et al.* Sensor fusion for the robust detection of facial regions of neonates using neural networks. *Sensors* 23, 4910 (2023).

25. Cao, Q., Shen, L., Xie, W., Parkhi, O. M. & Zisserman, A. VGGFace2: A dataset for recognising faces across pose and age. In *International Conference on Automatic Face and Gesture Recognition* (2018).

26. Liu, Z., Luo, P., Wang, X. & Tang, X. Deep learning face attributes in the wild. In *Proceedings of International Conference on Computer Vision (ICCV)* (2015).

27. Jain, V. & Learned-Miller, E. Fddb: A benchmark for face detection in unconstrained settings. Tech. Rep., UMass Amherst technical report (2010).

28. Yang, S., Luo, P., Loy, C. C. & Tang, X. Wider face: A face detection benchmark. In *IEEE Conference on Computer Vision and Pattern Recognition (CVPR)* (2016).

29. Cobo, M. M. *et al.* Multicentre, randomised controlled trial to investigate the effects of parental touch on relieving acute procedural pain in neonates (petal). *BMJ open* 12, e061841 (2022).

30. Hauck, A. G. *et al.* Effect of parental touch on relieving acute procedural pain in neonates and parental anxiety (petal): a multicentre, randomised controlled trial in the uk. *The Lancet Child & Adolesc. Heal.* 8, 259–269 (2024).

31. Stevens, B. J. *et al.* The premature infant pain profile-revised (pipp-r): initial validation and feasibility. *The Clin. journal pain* 30, 238–243 (2014).

32. Wada, K. Labelme: Image polygonal annotation with python. GitHub: https://github.com/wkentaro/labelme (2021).Version 5.0.1. DOI: 10.5281/zenodo.5711226.

33. Jocher, G. & Qiu, J. Ultralytics yolov11. GitHub: https://github.com/ultralytics/ultralytics (2024). Version 11.0.0.

34. Lin, T.-Y. *et al.* Microsoft coco: Common objects in context. In *Computer Vision–ECCV 2014: 13th European Conference, Zurich, Switzerland, September 6-12, 2014, Proceedings, Part V 13*, 740–755 (Springer, 2014).

35. PyTorch Team. Torchvision image transforms - pytorch documentation (2026). URL https://docs.pytorch. org/vision/main/transforms.html.

36. Zheng, Z. *et al.* Enhancing geometric factors in model learning and inference for object detection and instance segmentation. *IEEE Transactions on cybernetics* 52, 8574–8586 (2021).

37. Li, X. *et al.* Generalized focal loss: Towards efficient representation learning for dense object detection. *IEEE transactions on pattern analysis machine intelligence* 45, 3139–3153 (2022).

38. Kiefer, J. & Wolfowitz, J. Stochastic estimation of the maximum of a regression function. *The Annals Math. Stat.* 462–466 (1952).

# Supplementary Information

## InfantFace: Detecting infant faces in neonatal clinical environments

### Supplementary materials A: WIDER FACE category distribution

To better understand the composition of the WIDER FACE dataset, we visualised the number of images per category in the available dataset. Each category in the WIDER FACE dataset corresponds to a specific scene or context, such as 'parade' or 'sports'. Figure A.1 shows a histogram of all categories, with separate bars indicating the number of images in the released training and validation subsets.

### Supplementary materials B: Annotation inconsistencies in public datasets

Figure B.2 illustrates cases where the number of annotated faces did not match the actual number of visible faces in the images. These inconsistencies were observed across several public datasets, including the VGGFace2, CelebA, FDDB and WIDER FACE datasets; and motivated the development of our single-face verification process.

### Supplementary materials C: Convert FDDB annotation

The original FDDB annotations utilise elliptical regions rather than bounding boxes. To integrate these annotations into our training pipeline, we converted each ellipse into an axis-aligned bounding box that fully encompasses the ellipse. The bounding box width *w* and height *h* are computed as:

$$w = 2\sqrt{(a\cos\theta)^2 + (b\sin\theta)^2} \tag{C.1}$$

$$h = 2\sqrt{(a\sin\theta)^2 + (b\cos\theta)^2} \tag{C.2}$$

The bounding box centre remains the same as the ellipse centre:

$$x = X_0, \quad y = Y_0 \tag{C.3}$$

---

where $a$ and $b$ are the semi-major and semi-minor axes of the ellipse, respectively; $\vartheta$ is the rotation angle in radians; and $(X_0, Y_0)$ denotes the centre of the ellipse. Examples of the original and converted annotations are shown in figure C.3.

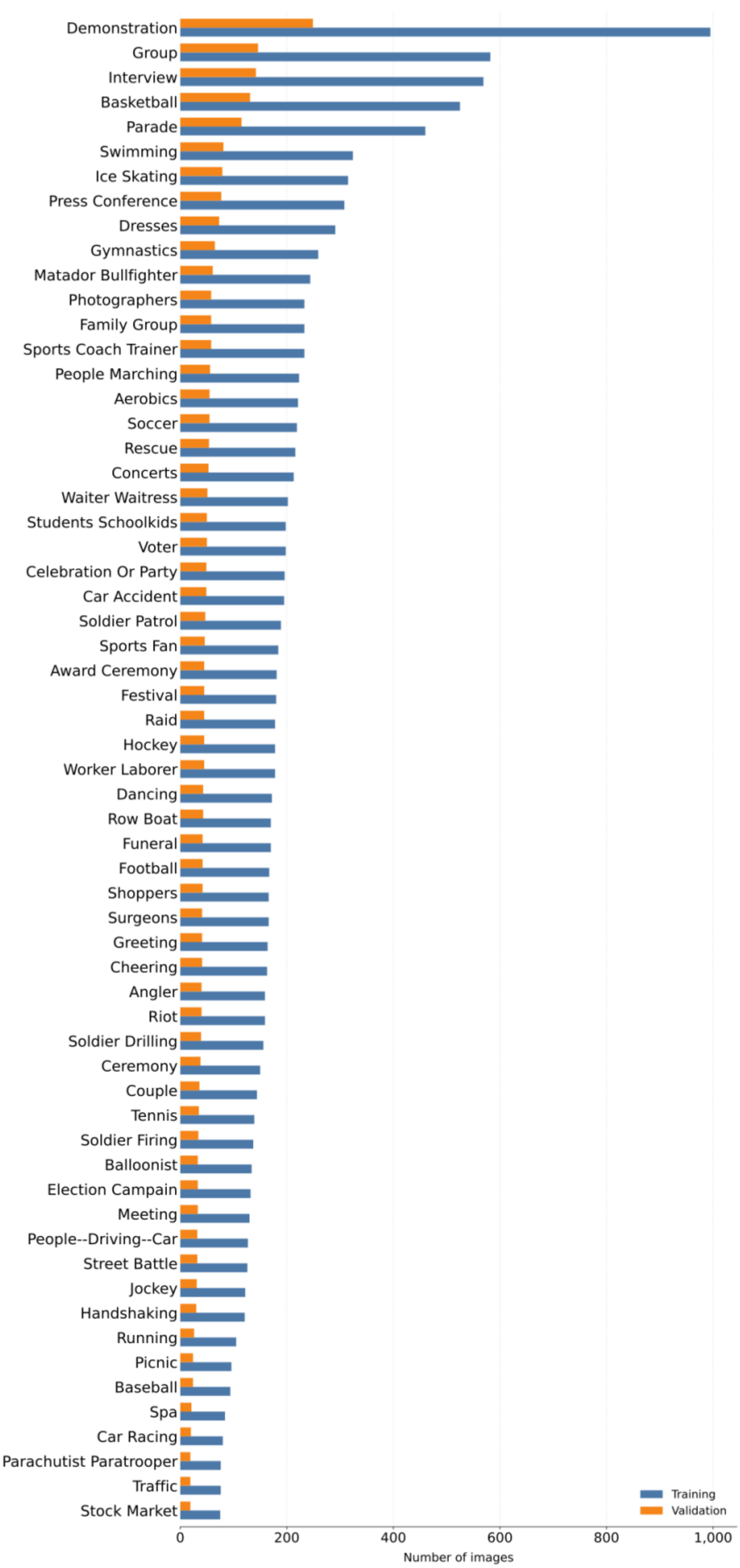

**Figure A.1.** Number of images per category in the publicly available WIDER FACE training and validation sets. Blue bars represent images in the original training set, and orange bars represent images in the validation set. Categories are sorted by total image count.

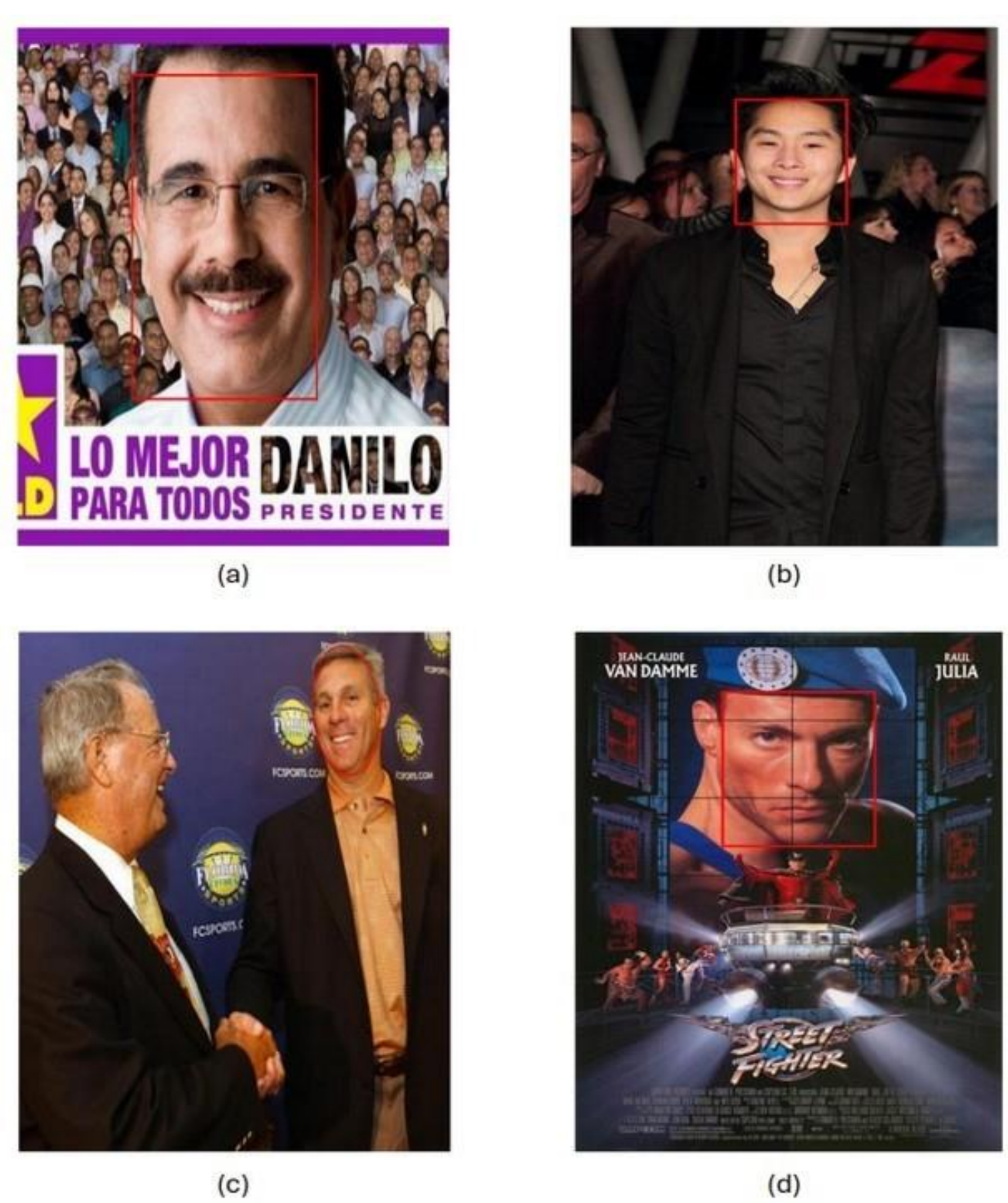

**Figure B.2.** Examples of official annotations for the four public datasets used in this study. (a) VGGFace2, (b) CelebA, (c) Face Detection Data Set and Benchmark (FDDB) and (d) WIDER FACE. Each image shows the ground truth face annotation provided by the dataset's authors. While all four datasets label these as single-face images, other visible but unannotated faces can be seen. The annotation formats differ: bounding boxes are used in (a), (b) and (d), while FDDB (c) provides elliptical annotations. Images have been slightly cropped to improve visual alignment and ensure consistency in presentation.

**Supplementary materials D: Face and image stats in VGGFace2**

Selecting the image size for training our InfantFace (pre-adaptation) model and the range for multi-scale training was a data-driven decision. We analysed two key properties of the VGGFace2 dataset: the face-to-image ratio and the distribution of image dimensions.

The face-to-image ratio, shown in figure D.4, demonstrates that most faces occupy around 20% of their source image area, with the distribution peaking near 0.2 and a small fraction of images have smaller face regions. We also reviewed the distribution of image dimensions, as illustrated in figure D.5, where the majority of images fall in the 100-299 pixel range, indicating that a target resolution of *128 × 128* pixels balances preserving visual detail and computational efficiency, given the large number of images.

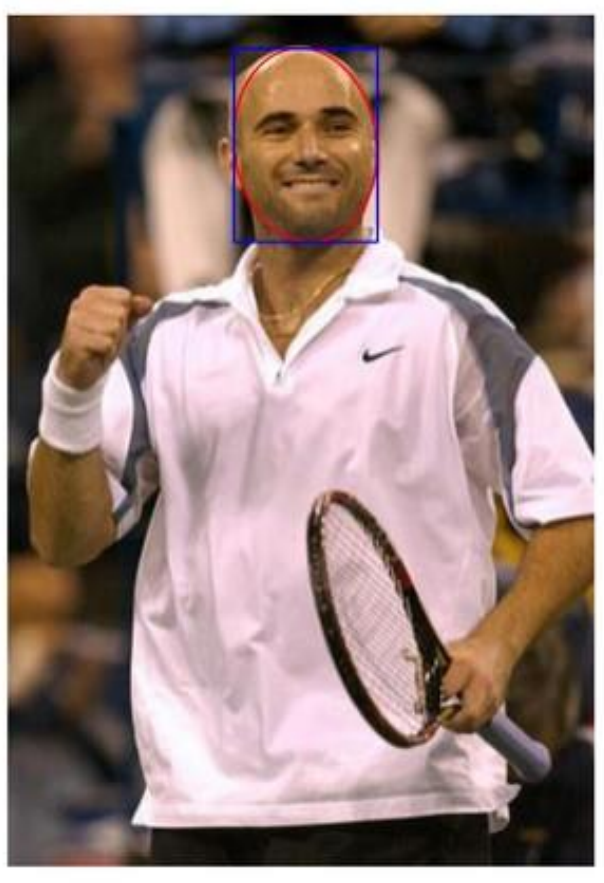
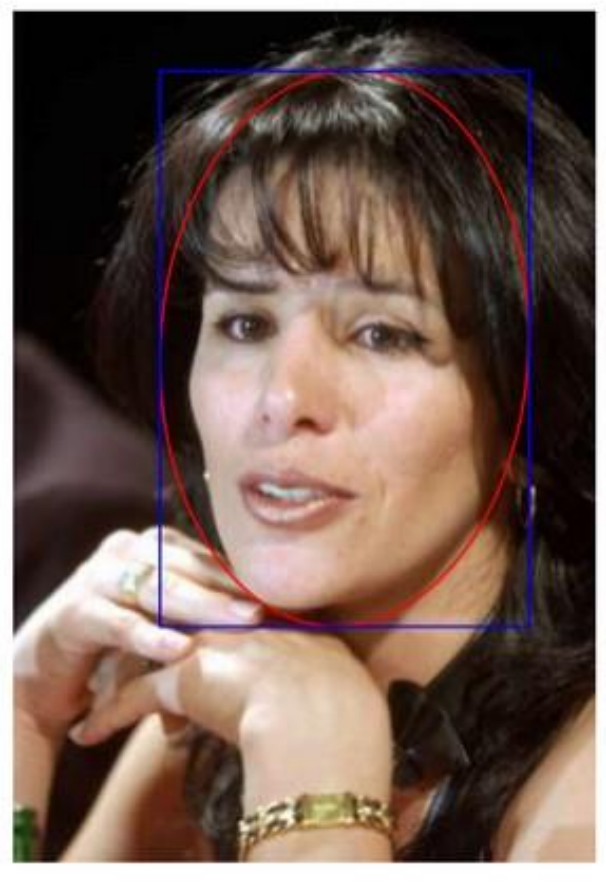
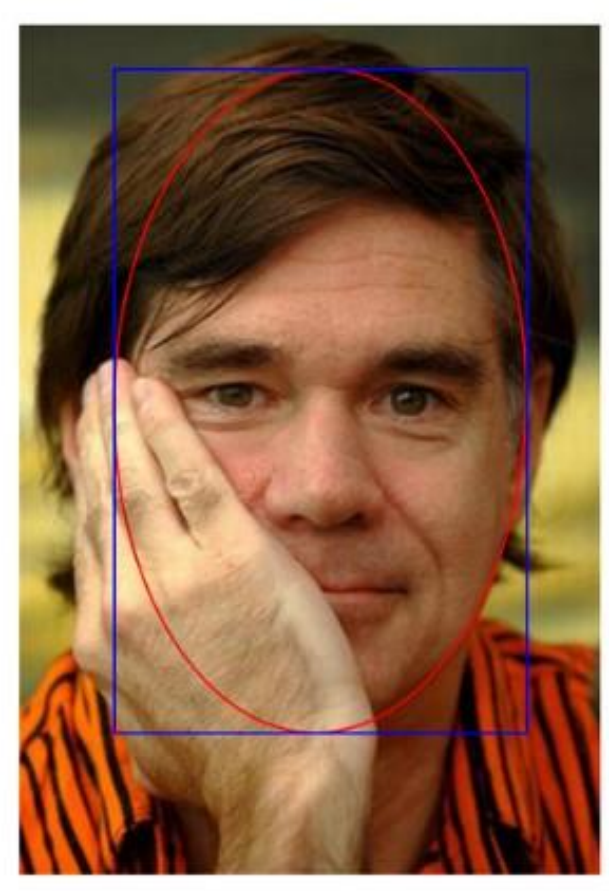

**Figure C.3.** Comparison of Face Detection Data Set and Benchmark (FDDB) annotation formats. Illustrative examples from the FDDB dataset show the original annotation format (red ellipse) and the corresponding converted format (blue rectangle), which is widely adopted in modern face detection benchmark datasets.

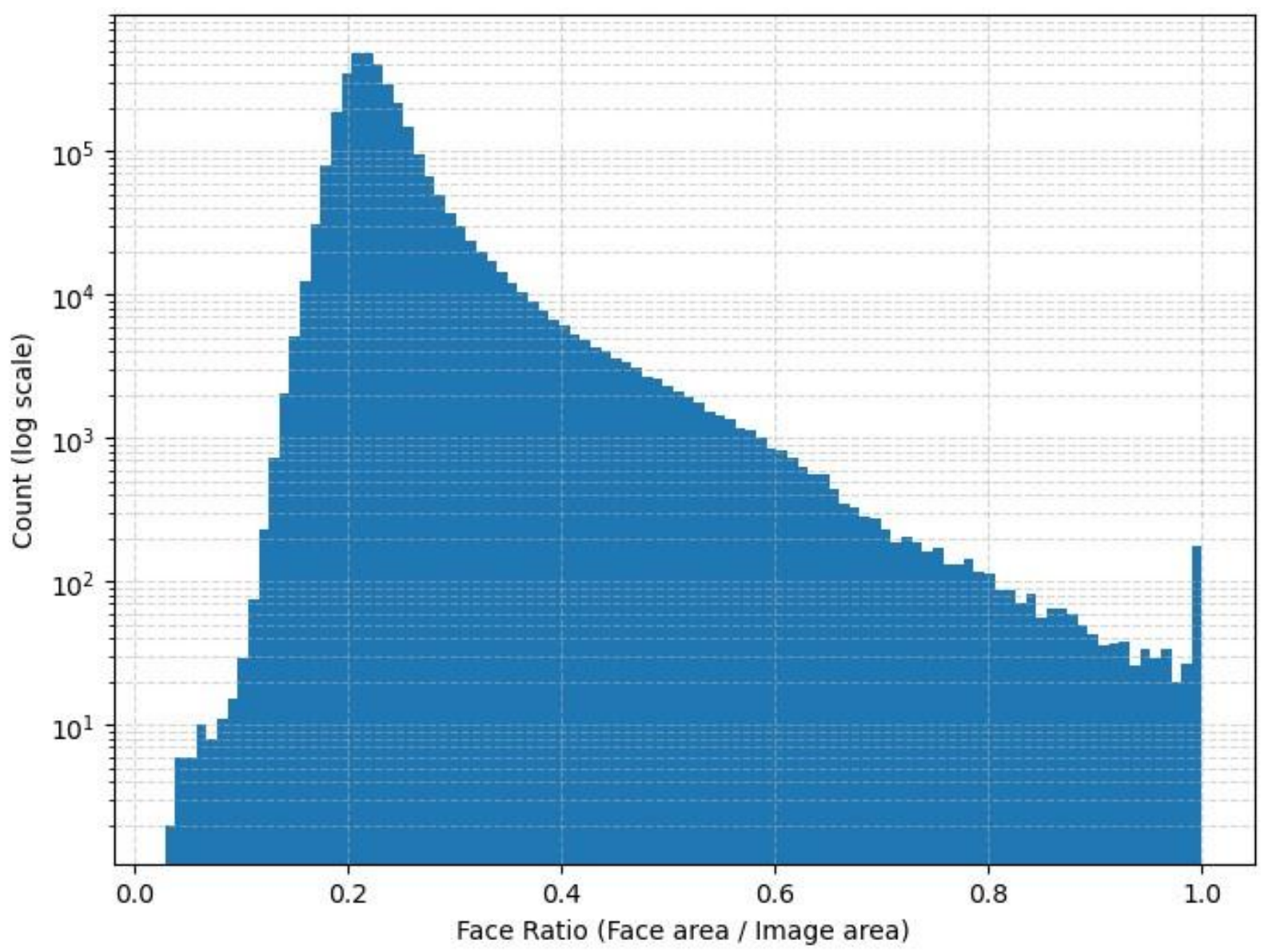

**Figure D.4.** Face-to-image ratio in VGGFace2 dataset.

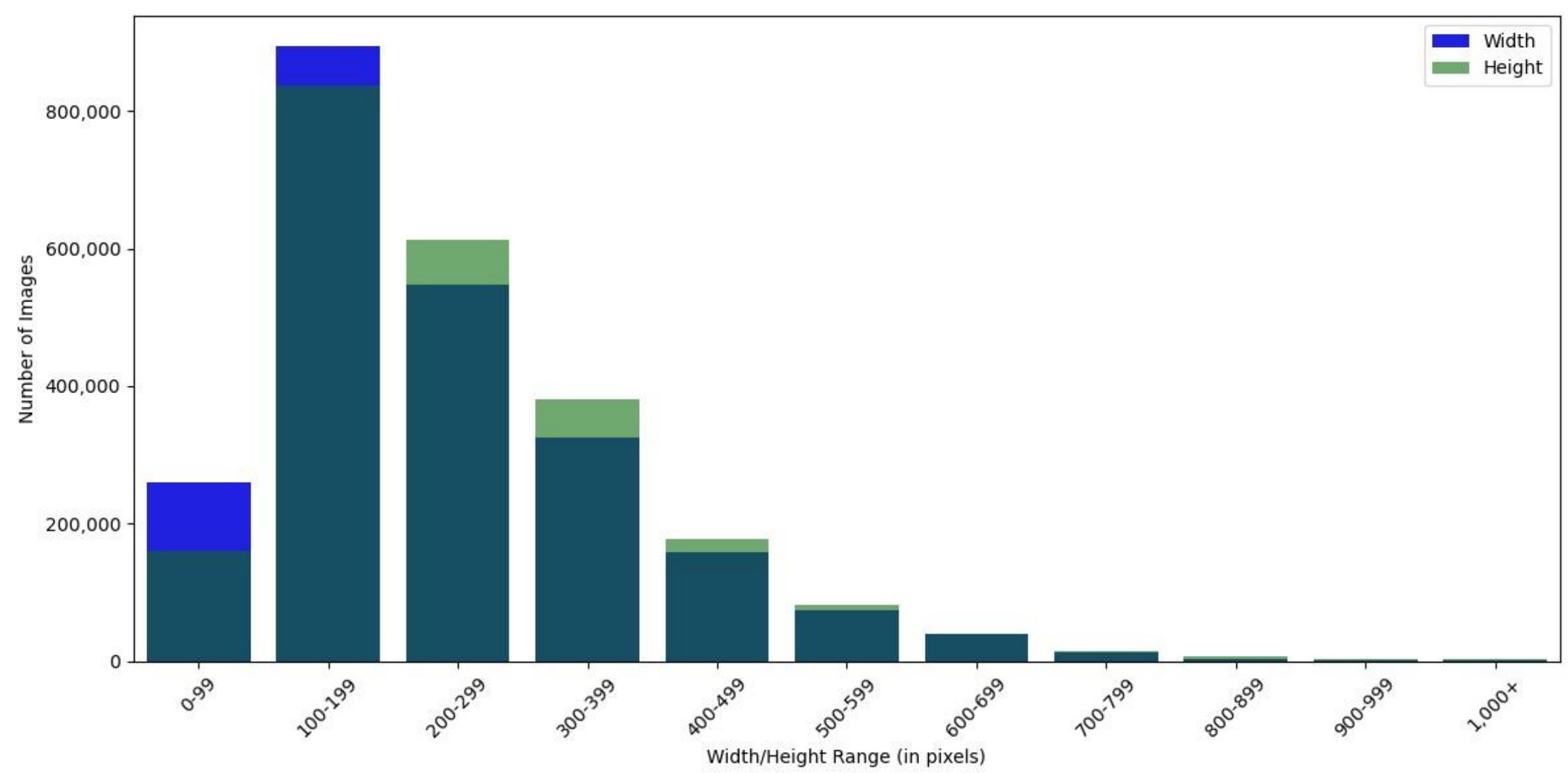

**Figure D.5.** Distribution of image dimensions (width and height) in the VGGFace2 dataset, grouped by pixel ranges.